\documentclass[10pt]{cai}
\usepackage{hyperref}

\graphicspath{{figs/}}

\newcommand{\camembertFrozen}{\textbf{CamemBERT frozen}}
\newcommand{\camembertUnfrozen}{\textbf{CamemBERT unfrozen}}
\newcommand{\camembertUnfrozenWarmup}{\textbf{CamemBERT unfrozen warmup}}
\newcommand{\biLstm}{\textbf{bi-LSTM}}
\newcommand{\guillemet}[1]{``#1''}

\definecolor{blue}{HTML}{636EFA}
\definecolor{rouge}{HTML}{EF553B}
\definecolor{vert}{HTML}{00CC96}
\definecolor{mauve}{HTML}{AB63FA}
\definecolor{orange}{HTML}{FFA15A}

\definecolor{skyblue}{HTML}{19D3F3}
\definecolor{sulu}{HTML}{B6E880}
\definecolor{pink}{HTML}{FF6692}
\definecolor{violet}{HTML}{FF97FF}

\sethlcolor{lichen} %for soul pacakge

\newcommand{\blueemph}[1]{\textcolor{blue}{\textbf{#1}}}
\newcommand{\rougeemph}[1]{\textcolor{rouge}{\textbf{#1}}}
\newcommand{\vertemph}[1]{\textcolor{vert}{\textbf{#1}}}
\newcommand{\mauveemph}[1]{\textcolor{mauve}{\textbf{#1}}}
\newcommand{\orangeemph}[1]{\textcolor{orange}{\textbf{#1}}}

\newcommand{\skyblueemph}[1]{\textcolor{skyblue}{\textbf{#1}}}
\newcommand{\sulueemph}[1]{\textcolor{sulu}{\textbf{#1}}}
\newcommand{\pinkemph}[1]{\textcolor{pink}{\textbf{#1}}}
\newcommand{\violetemph}[1]{\textcolor{violet}{\textbf{#1}}}

\newcommand\blfootnote[1]{%
  \begingroup
  \renewcommand\thefootnote{}\footnote{$^\diamond$#1}%
  \addtocounter{footnote}{-1}%
  \endgroup
}

\begin{document}
\begin{center}

  \title{\guillemet{FIJO}: a French Insurance Soft Skill Detection Dataset}
  \maketitle

  \thispagestyle{empty}

\begin{tabular}{cc}
    David Beauchemin\upstairs{\affilone}$^\diamond$, Julien Laumonier\upstairs{*\affiltwo}$^\diamond$, Yvan Le Ster\upstairs{\affilone}$^\diamond$, Marouane Yassine\upstairs{\affiltwo}$^\diamond$
  \\[0.25ex]
  {\small \upstairs{\affilone} Departement of Computer Science and Software Engineering, Université Laval, Québec, Canada} \\
  {\small \upstairs{\affiltwo} Institute Intelligence and Data, Université Laval, Québec, Canada}
\end{tabular}
  
\emails{\upstairs{*}julien.laumonier@iid.ulaval.ca}
\vspace*{0.2in}
\end{center}

\begin{abstract}
Understanding the evolution of job requirements is becoming more important for workers, companies and public organizations to follow the fast transformation of the employment market. Fortunately, recent natural language processing (NLP) approaches allow for the development of methods to automatically extract information from job ads and recognize skills more precisely. However, these efficient approaches need a large amount of annotated data from the studied domain which is difficult to access, mainly due to intellectual property. This article proposes a new public dataset, FIJO, containing insurance job offers, including many soft skill annotations. To understand the potential of this dataset, we detail some characteristics and some limitations. Then, we present the results of skill detection algorithms using a named entity recognition approach and show that transformers-based models have good token-wise performances on this dataset. Lastly, we analyze some errors made by our best model to emphasize the difficulties that may arise when applying NLP approaches.
\end{abstract}

\copyrightnotice

\section{Introduction}
\blfootnote{Authors contributed equally to this work.}
The digital transformation's impact on professional practices has led to a rapid evolution of in-demand job skills, making it difficult to track these changes for enterprises and workers. Manual evaluation is becoming exceedingly complex and time-consuming, justifying the need for an automatic evaluation of these changes \cite{bakhshi_future_2017}. One way to study these changes in job ads is automatic skills recognition \cite{squicciarini_demand_2021}.

However, job offer data is not easy to access even in job offers web platforms, mainly due to intellectual property issues. Moreover, the annotation needed to achieve good skill recognition performances through supervised machine learning is another complex and costly task. Indeed, many datasets offering job descriptions are accessible online such as the \verb|mycareersfuture| public dataset \cite{bhola_retrieving_2020}. However, as \cite{khaouja2021survey} reported in their article, very few public annotated datasets exist, and none are in French.

As contributions, in this article, we propose \guillemet{\href{https://github.com/iid-ulaval/FIJO-dataset}{\color{red}French Insurance Job Offer (FIJO)}}, a free and public non-annotated and annotated dataset, to facilitate research in this domain. This dataset focuses on soft skills, which describe the way employees work alone and with others, instead of hard skills, which represent a more formal knowledge used at work \cite{lyu2021soft}. Also, we explore the training of a token-wise NER French skill detection algorithm in the field of insurance with state-of-the-art algorithms\footnote{All code used to obtain these results will be available on GitHub \href{https://github.com/iid-ulaval/FIJO-code}{\color{red}here}.}. 

The rest of this paper is structured as follows. Firstly, we begin with a brief overview of the literature on skill detection in \autoref{sec:relatedwork}. Secondly, we will present FIJO in \autoref{sec:data}, including how we constructed our French complexity corpus, some statistics and analysis of the dataset. Third, we will present our skill detection algorithm, the training settings and our results in \autoref{sec:skilldetection}. Finally, we will draw some concluding remarks in \autoref{sec:conclusion}.

\section{Related work}
\label{sec:relatedwork}

The first approach to recognizing skills inside a job offer is using statistical techniques. It is a commonly used approach in many pieces of work such as \cite{malherbe_bridge_2016} who detect hard and soft skills in job offers by matching a list of keywords in the text with a skill database. Their skill database uses external databases knowledge, namely DBPedia and StackOverflow. Other works \cite{sodhi_content_2010, gardiner_skill_2018, squicciarini_demand_2021} do not rely on skills databases but use content analysis to detect the presence of certain words or concepts in offers.

Other approaches using machine learning to detect skills in job offers have received lots of attention recently \cite{khaouja2021survey}. The skill recognition problem has been modeled either using topic modeling through Latent Dirichlet Analysis \cite{gurcan_big_2019}, text classification with CNN and LSTM \cite{sayfullina2018learning}, or NER with LSTM \cite{jia_representation_2018} and transformer-based models \cite{tamburri_dataops_2020}.
However, these pieces of work mainly focus on specialized skills (e.~g. IT skill \cite{gurcan_big_2019}), focus on soft skill \cite{sayfullina2018learning} or are applied on English job ads only.

As per \cite{khaouja2021survey}, the conclusion in their survey on skill identification shows that very few datasets are available online.
Most of the recent works do not release their dataset nor mention the reasons for the non-publication.
For example, \cite{cerioli20205} uses content analysis, and 5 million non-annotated jobs adds to determine whether or not testing software is a standard in the IT industries.
The possible reason for the lack of publication by the authors was possibly due to the intellectual property constraint from their industrial partner.
However, a recent new public dataset released by \cite{bhola_retrieving_2020} focuses on extracting identified hard skills in job ads. 
The dataset consists of 20,298 job ads, where each ad includes nearly 20 hard skills on average.
A unique skill term corresponds to a unique class. Thus, the overall dataset includes 2,548 skill classes. 
For example, \guillemet{Microsoft Word} is a skill, and \guillemet{Microsoft Excel} is another skill.
This new dataset lacks annotation for soft skills that have been more required than hard skills by enterprises in the past decade \cite{lyu2021soft}.
\vspace{-1em}
\section{Data}
\label{sec:data}

FIJO was created in partnership with four Canadian insurance companies. 
The dataset consists of non-annotated and annotated French job ads published by them, as well as their metadata (e. g. date of publication) between the years 2009 to 2020.
Each job offer's text was manually extracted and semi-manually cleaned using the following procedure: removal of carriage return in an incomplete sentence (this is due to bullet point text) and multiple carriage returns, removal of bullet point character, normalization of the apostrophe punctuation characters, and removing of trailing whitespace at the beginning or end of a sentence.
In order to protect the interests of the companies to whom the published data belongs, we chose to de-identify the job ads before making them publicly available. This process consists of three steps. 
Firstly, we used regular expressions to substitute the different variations of the companies' names and email addresses present in the offers. 
Next, a SpaCy French pre-trained NER model (\verb|fr_core_news_lg|) was used to identify potential names and locations to help with the following step. 
Finally, a manual check was conducted on each offer to substitute the following elements: names, locations, postal addresses and miscellaneous elements that could help identify the companies (i. e. products, department names). \autoref{tab:substitution tags} describes the substitution tags employed.

\begin{table}
    \centering
        \begin{tabular}{lc}
        \toprule
             Substitution tag & Description \\ \midrule
             <anon\_name> & A person's name \\
             <anon\_location> & A postal address or a city \\
             <anon\_company> & One of the companies' names \\
             <anon\_misc> & An element that can help identify one of the companies \\ \bottomrule
        \end{tabular}
        \caption{Substitution tags used for de-identification}
        \label{tab:substitution tags}
\end{table}
\vspace{-1em}
\subsection{Dataset Statistics}

\begin{figure}
    \centering
    \begin{minipage}{.4\textwidth}
      \centering
        \captionsetup{width=.9\linewidth}
        \includegraphics[width=\linewidth]{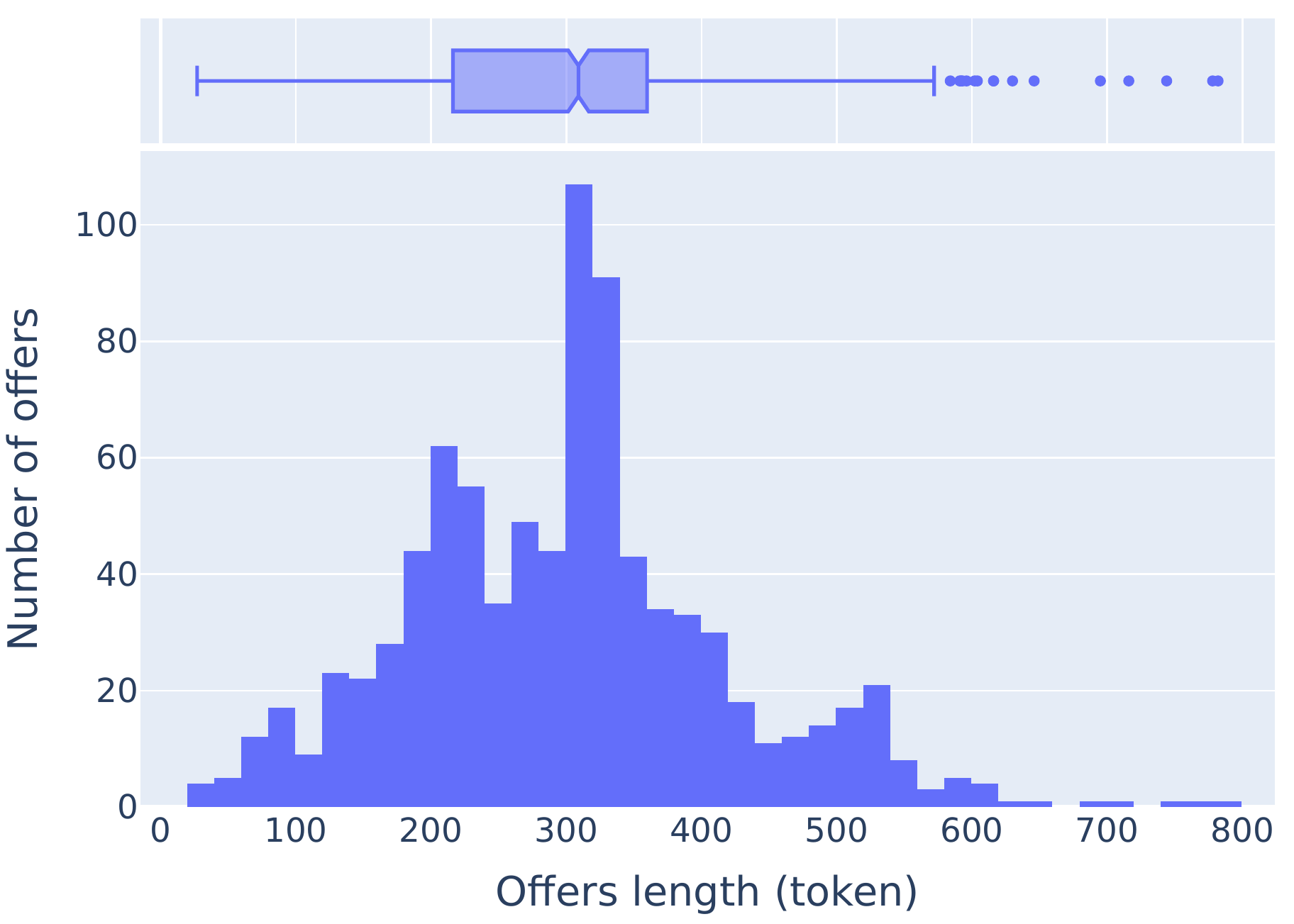}
        \caption{Ads length by words}
        \label{fig:histofferlen}
    \end{minipage}%
    \begin{minipage}{.6\textwidth}
        \centering
        \captionsetup{width=.9\linewidth}
        \includegraphics[width=\linewidth]{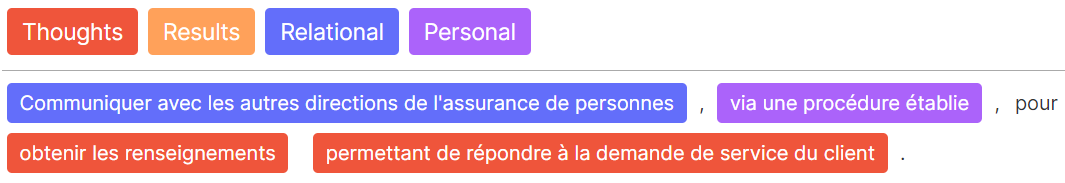}
        \caption{Example of a French annotation where each color refers to a class (i. e. skill)}
        \label{fig:annotatonexample}
    \end{minipage}
\end{figure}

The dataset is composed of 867 de-identified French job ads. 
As shown in \autoref{fig:histofferlen}, job ads lengths vary greatly, with an average length of 300.97 and a standard deviation of 119.78 tokens\footnote{Punctuations are computed as a token. We do so since our pre-processing procedure does not include the removal of punctuation nor lemmatization, or stemming.}. 
We can also observe that a few offers (16) are outliers with a length of more than 572.
Moreover, \autoref{tab:nonannotatedstatistics} presents statistics of the dataset, where the lexical richness corresponds to the ratio of a job offer's number of unique words over the vocabulary cardinality without removing the stop words or normalizing them \cite{van2007comparing}.
We can see that the lexical richness is relatively low, which means they are quite similar in terms of vocabulary.

\begin{table}
    \centering
        \begin{tabular}{lclclc}
        \toprule
        Average \# of Words & 300.97     &  \# of Words & 260,942 & Average \# of Sentences & 20.66\\
         \# of Unique Words & 5,931 & Average Sentence Length & 14.57 & Average Lexical Richness & 0.023\\ 
        \bottomrule
        \end{tabular}
        \caption{Unannotated dataset statistics} \label{tab:nonannotatedstatistics}
\end{table}

\subsection{Annotated Dataset}
To learn to identify soft skills inside ads, 47 offers were annotated, and more precisely, each job offer sentence, for a total of 499 annotations.
Our annotation process consists of creating a skills reference, which defines the skills used for the annotations, and randomly selecting 47 offers to be annotated by a domain expert.
Annotation was conducted with non-overlapping sentence entities done individually. However, the overall job offer was given as a reference to the annotator for context. 
Each entity contains at least one word or, at most, the complete sentence.

Based on the skill groups of the \href{http://catalogue.iugm.qc.ca/GED_IUG/109311392759/Referentielcomp.pdf}{\textit{AQESSS}} public skills repositories and the one used by our insurance partners, which are based on the commercial \href{https://www.kornferry.com/}{Korn Ferry} and \href{https://humance.ca/}{SPB} repositories, a set of four skills have been identified. 
Namely, \guillemet{Thoughts}, \guillemet{Results}, \guillemet{Relational} and \guillemet{Personal}\footnote{The tag are written in French in the dataset. Namely, \guillemet{Pensée}, \guillemet{Résultats}, \guillemet{Relationnel} and \guillemet{Personnel}.}. 
The number of classes has been limited to four mainly because, in general, learning algorithms are not known for being inefficient on a large number of tags \cite{JMLR:v15:gupta14a} but also because of the possible confusion between skills during the annotation process (see \autoref{sec:limitations}). \autoref{fig:annotatonexample} presents an example of a sentence annotation.

\subsection{Annotated Dataset Statistics}
\begin{figure}
     \centering
    \begin{minipage}{.45\textwidth}
        \centering
        \captionsetup{width=.9\linewidth}
            \begin{tabular}{lclc}
                \toprule
                Av. \# of & & \# of &\\\midrule
                Words & 270.26     & Words & 12,702 \\
                Sentences & 16.28 & Unique Words & 1,902\\\bottomrule
            \end{tabular}
            \captionof{table}[foo]{Annotated dataset statistics} 
            \label{tab:annotatedstatistics}
     \end{minipage}
    \begin{minipage}{.49\textwidth}
      \centering
        \captionsetup{width=.9\linewidth}
        \includegraphics[width=\textwidth]{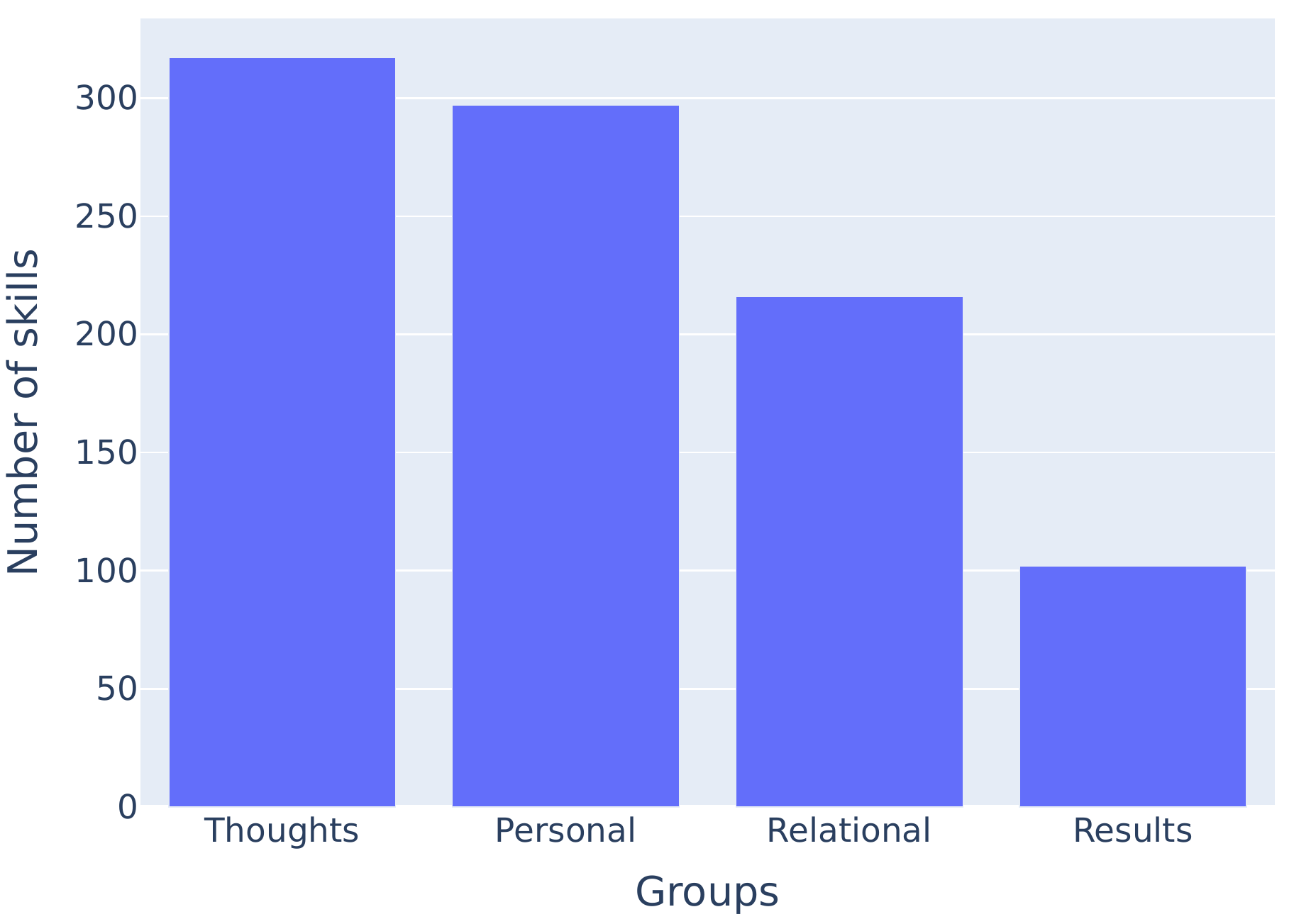}
         \caption{Number of entities per class}
         \label{fig:ent_nb}
     \end{minipage}
     
    \begin{minipage}{.49\textwidth}
      \centering
        \captionsetup{width=.9\linewidth}
         \centering
        \includegraphics[width=\textwidth]{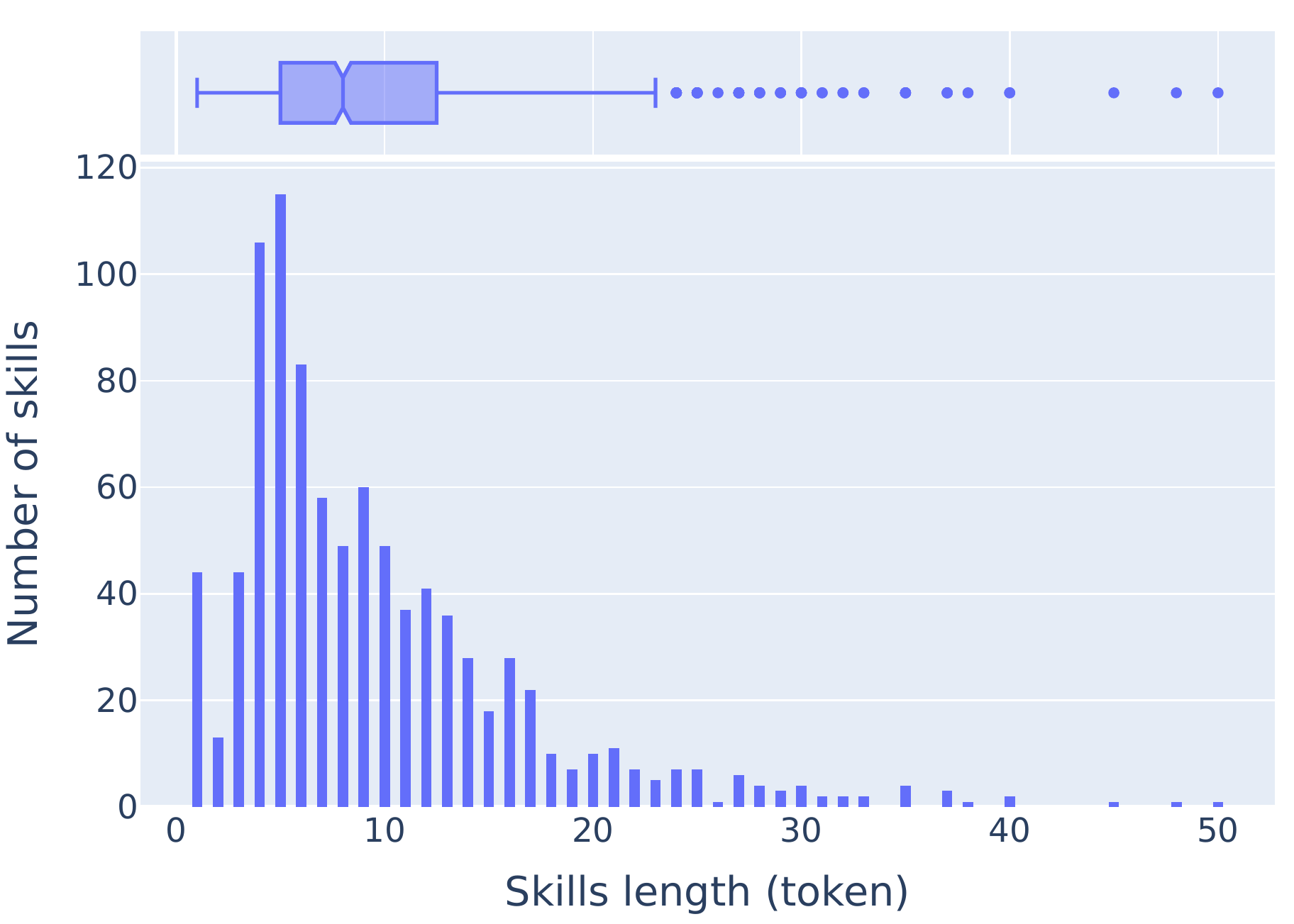}
         \caption{Distribution of entities length base on their tokens}
         \label{fig:ent_len}
     \end{minipage}
    \begin{minipage}{0.45\textwidth}
      \centering
        \captionsetup{width=\linewidth}
         \centering
        \includegraphics[width=\textwidth]{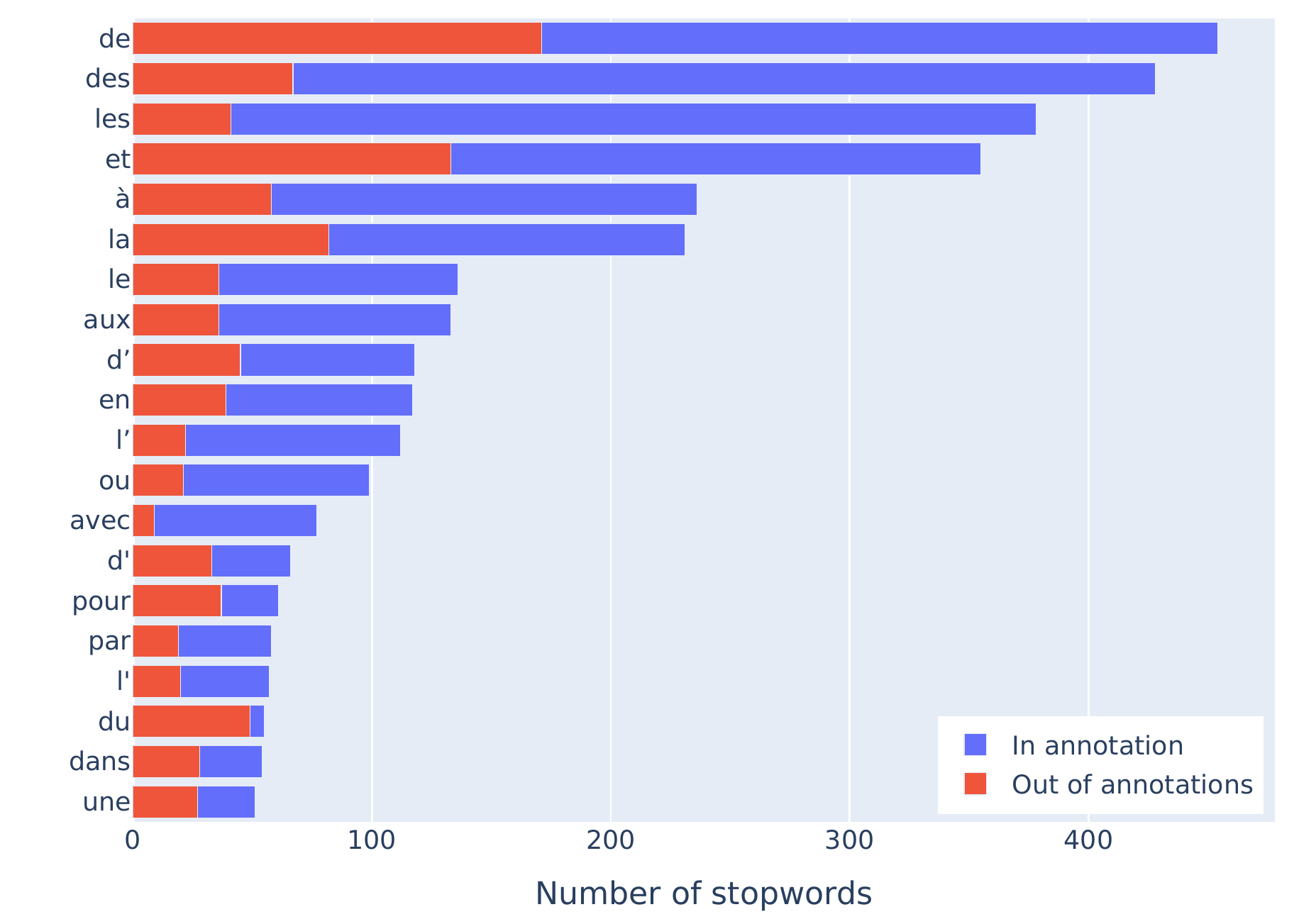}
         \caption{Number of time a stopwords occur in an entity (\blueemph{blue}) or out  of an entity (\rougeemph{red})}
         \label{fig:stopwords}
     \end{minipage}
    \vspace{-1em}
\end{figure}

First, as shown in \autoref{fig:ent_nb} our annotated portion of FIJO consists of 932 entities distributed unevenly between the four classes.
We can see that the class with the most entities is \guillemet{Thoughts} with 317 entities, followed by \guillemet{Personal} with 297 and \guillemet{Relational} with 216.
The class with the lowest number of entities is \guillemet{Results} with 102 entities. Second, as illustrated in \autoref{fig:ent_len}, our entities are on average 9.6 long with a standard deviation of 7.14 tokens.
Their length ranges from a single token to 50 tokens, but 50\% are below 8.
Moreover, Table~\ref{tab:annotatedstatistics} presents statistics of the dataset.
We have a similar average number of words and sentences as per the non-annotated dataset.
However, our annotated dataset uses fewer words and even fewer unique words.
Finally, \autoref{fig:stopwords} presents the number of occurrences of stop words in an entity text (\blueemph{blue}) or outside of an entity (\rougeemph{red}).
It shows that some stop words are overly represented in entities, such as \guillemet{\textit{de}} and \guillemet{\textit{des}} that are mostly in annotations since long skills tend to contain a high number of stop words.

\vspace{-0.5em}
\subsection{Dataset Limitations}
\label{sec:limitations}
We have identified three limitations to our dataset: unbalanced entities classes, lexical overlapping and soft skill identification.

Firstly, as illustrated in \autoref{fig:ent_nb}, our annotated dataset is composed of an unbalanced number of classes where two classes are more represented than the two others.
When classes imbalance exists in training data, a classification algorithm will typically over-classify the majority group (\guillemet{Thoughts}) due to its increased prior probability. As a result, the instances belonging to the minority group (\guillemet{Results}) will likely be misclassified more often than those belonging to the majority group \cite{johnson2019survey}.

\begin{figure}
    \centering
    \captionsetup{width=\linewidth}
    \includegraphics[width=0.6\textwidth]{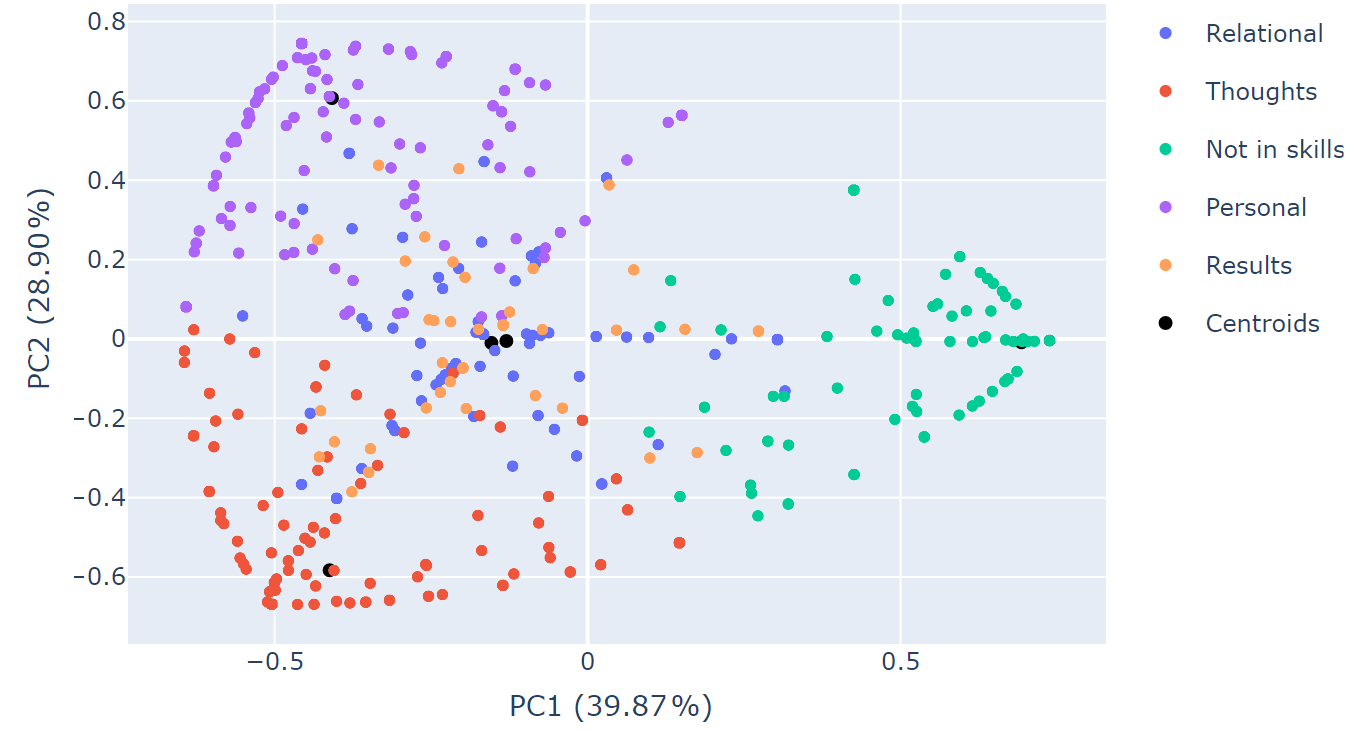}
    \caption{2-dimensions\protect\footnotemark{} PCA of the TF-IDF separated by entities groups with the removal of stop words and lemmatization of terms}
    \label{fig:pca_tf-idf}
\end{figure}

Secondly, \autoref{fig:pca_tf-idf} illustrates the 2-dimension PCA of the TF-IDF score of each entity's text after stop words removal and lemmatization, separated by class.
It shows that we can separate the terms (and centroids) present in skills from those that are not.
For example, the \guillemet{Personal} (purple) centroid (upper left) and \guillemet{Thoughts} (red) centroid (down left) can be distinctly separated from each other and the other two centroids.
Such separation may make it easier to discern the two cases since skills use specific terms that are less common in other skill texts.
For example, the word \guillemet{\textit{collaborer}} (collaborate) appears only in the \guillemet{Personal} entities.
By contrast, well-distributed terms among the different groups may be more challenging.
For example, the word \guillemet{\textit{atteindre}} (achieve) occurs in all four classes.
However, we can also see that some terms are quite close to each other, possibly leading to a more difficult distinction between the four classes' word terminologies.

\footnotetext{The third dimension is not represented here due to difficulty representing the third dimension in a non-reactive figure. However, we would like to clarify that using the third dimension makes the \guillemet{Relational} and \guillemet{Results} class way more spaced out than in two dimensions only.}

\begin{figure}
    \begin{minipage}{\textwidth}
        \centering
        \captionsetup{width=\linewidth}
        \includegraphics[width=0.55\textwidth]{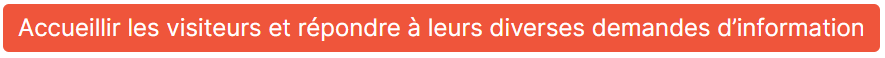}
         \caption{An example of a possible confusion between the classes \guillemet{Thoughts} and \guillemet{Relational} (\guillemet{Welcoming visitors and responding to their various requests for information})}
         \label{fig:anot_conf1}
    \end{minipage}

    \begin{minipage}{.48\textwidth}
      \centering
        \captionsetup{width=\linewidth}
         \centering
        \includegraphics[width=\textwidth]{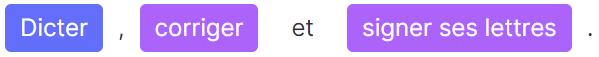}
         \caption{An example of two consecutive skill annotations with the same class that are separated by a coordinating conjunction}
         \label{fig:anot_conf2}
    \end{minipage}
    \hspace{1pt}
    \begin{minipage}{0.48\textwidth}
      \centering
        \captionsetup{width=\linewidth}
         \centering
        \includegraphics[width=\textwidth]{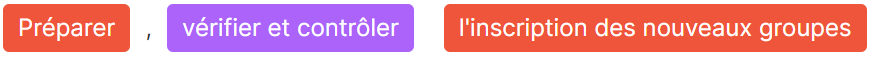}
         \caption{An example of an entity \guillemet{squeezed} between two other entity of a different class}
         \label{fig:anot_conf3}
     \end{minipage}
    \vspace{-1em}
\end{figure}

Finally, soft skill identification is not an easy task, as mentioned by \cite{squicciarini_demand_2021}, and our dataset reflects it.
Firstly, some distinctions between skills can be quite confusing, as seen in the example in \autoref{fig:anot_conf1}.
This example can be read as \guillemet{Welcoming visitors and responding to their various requests for information} and is tagged as Thoughts.
However, one reader might find that such a skill could also represent a Relational one. 
Thus, creating a confusing distinction between some examples.
Secondly, some examples contain two consecutive skills of the same class separated by coordinating conjunctions as seen in \autoref{fig:anot_conf2}.
We can see that the French coordinating conjunction \guillemet{\textit{et}} (and) split the two Personal entities.
However, such coordinating conjunction is not always used to do so. It can also be used as an addition of information, such as in \guillemet{\textit{vérifier et contrôler}} (verify and control).
Thus, it can be quite challenging to determine if a subset of a sentence is one or two skills in some cases.
Finally, it is common to see job ads that list expected soft skills in the same sentence, but all skills do not belong to the same skill class.
An example of such switching between two entity types is illustrated in \autoref{fig:anot_conf3}. 
We can see that the first token is an entity, followed by another entity of a different type and the rest of the sentence is another entity of the same class as the first entity.
This kind of \guillemet{squeezing} of two entities sharing the same class around an entity of a different class can be challenging for a NER model. 
All these limitations justify the fact that, in this article, we apply token-wise approaches to the dataset to start with an easier learning task.
         
\section{Skill detection}
\label{sec:skilldetection}

Since skill detection is a sequence classification task similar to NER, we approach it using a recurrent neural network, namely a bidirectional long short-term memory (bi-LSTM) network \cite{lstm}. First of all, we encode each word in a given sequence using FastText's pre-trained French embedding model \cite{Bojanowski2017EnrichingWV} which produces 300-dimensional word embeddings. Once a sequence is encoded, we feed each of the word embeddings to our \biLstm{} which has a hidden state dimension of 300 and obtain a new representation for each word. The final step in the prediction process is to classify each word using a fully connected network comprised of one linear layer followed by a softmax activation function.

LSTM-based classifiers have proven to be quite efficient in terms of performance for sequence classification tasks. However, these models usually require a large amount of data to guarantee such performances. Therefore, since our annotated dataset contains a limited amount of data, we also use a pre-trained transformer model \cite{transformer} in a transfer learning setting. Our model of choice is CamemBERT \cite{martin-etal-2020-camembert} a French transformer encoder based on the BERT architecture \cite{DBLP:conf/naacl/DevlinCLT19}. Consequently, we use it to encode our text sequence, and we employ a fully connected network identical to the one employed with the \biLstm{} model in order to accomplish the classification. We experiment with two configurations of this model, one in which CamemBERT's weights are frozen and one in which they are not. We dub these models \camembertFrozen{} and \camembertUnfrozen{} respectively.

To further investigate the sensibility of our models to the amount of training data, as well as the transfer learning potential of the pre-trained transformer model, we experiment with different data subset sizes and report the results in \autoref{sec:results}.

\subsection{Experiments}

We train each of the aforementioned models five times using different random initialization seeds ($[5, 10, 15, 20, 25]$). The models were trained for 300 epochs at most with an initial learning rate of $0.01$ for \biLstm{} and \camembertFrozen{} and of $0.0001$ for \camembertUnfrozen{} as suggested by \cite{DBLP:conf/naacl/DevlinCLT19}. Therefore, a learning rate schedule that decreased the learning rate by a factor of 0.1 after every five epochs without any decrease of the validation cross-entropy loss was applied. An early stopping with a patience of 15 epochs to prevent overfitting. Additionally, for the \camembertUnfrozen{} model, we experimented with the training procedure and hyperparameters proposed in \cite{mosbach2021on} in order to address the possible training instability associated with fine-tuning transformer-based language models. As such, an additional five experiments (using the same random seeds) were run with \camembertUnfrozen{} by limiting the number of epochs to $20$ and scheduling the learning rate as follows: we start the training with a linear learning rate warmup (i. e. the learning rate was linearly increased) up to $0.2\mathrm{e}{-5}$ for the first $10\%$ of epochs, followed by a linear learning rate decay for the rest of the training epochs. We use \camembertUnfrozenWarmup{} to refer to this model.

The training data was divided using a $80\%-10\%-10\%$ train-validation-test split with simple random sampling, resulting in a total of $400$ training samples. We also experiment with different training data subsets. Each subset is composed by sampling the first X data samples from the full training set placed in order, with $\mathrm{X} \in \{50, 100, 150, 200, 350, 400\}$.

The models and training procedures were implemented using Poutyne \cite{poutyne}, HuggingFace's Transformers \cite{wolf-etal-2020-transformers} and spaCy \cite{Honnibal_spaCy_Industrial-strength_Natural_2020}.

\subsection{Results}
\label{sec:results}

\begin{table}
    \centering
    \captionsetup{width=\linewidth}
    \resizebox{\textwidth}{!}{\begin{tabular}{ccccc}
    \toprule
    Data subset size &                       \camembertUnfrozenWarmup &            \camembertUnfrozen &              \camembertFrozen &                          \biLstm \\
    \midrule
                50 &         $46.40 \pm 3.72$ &          $61.25 \pm 5.23$ &          $46.55 \pm 0.59$ &           $25.53 \pm 0.58$ \\
               100 &         $66.71 \pm 6.86$ &          $75.24 \pm 2.13$ &          $56.02 \pm 0.22$ &         $43.74 \pm 15.64$ \\
               150 &          $75.73 \pm 2.39$ &          $77.67 \pm 1.54$ &          $59.87 \pm 0.28$ &          $47.92 \pm 7.72$ \\
               200 &         $76.40 \pm 2.57$ &          $79.78 \pm 1.12$ &          $64.23 \pm 0.21$ &          $55.44 \pm 6.43$ \\
               250 &         $78.97 \pm 2.03$ &          $80.23 \pm 3.82$ &          $65.78 \pm 1.64$ &         $57.71 \pm 10.65$ \\
               300 &         $78.18 \pm 0.78$ &          $74.99 \pm 4.76$ &          $61.99 \pm 0.22$ &           $53.24 \pm 2.85$ \\
               350 &         $78.27 \pm 3.73$ &          $78.72 \pm 2.42$ & $\mathbf{67.47 \pm 0.21}$ &          $56.34 \pm 4.25$ \\
      400 & $\mathbf{80.85 \pm 1.67}$ & $\mathbf{83.69 \pm 1.80}$ &          $67.29 \pm 0.23$ & $\mathbf{60.69 \pm 9.23}$ \\
    \bottomrule
    \end{tabular}}
    \caption{Mean token-wise accuracy and one standard deviation on test data across different seeds for training subsets (bold values correspond to maximum mean accuracy for a model)}
    \label{tab:results}
    \vspace{-1em}
\end{table}

\autoref{tab:results} presents the mean token-wise accuracy and one standard deviation on the test set for the four trained models where bold values correspond to maximum mean accuracy for each model. As expected, fine-tuning the complete CamemBERT model yields the best performance with both \camembertUnfrozen{} and \camembertUnfrozenWarmup{} leading the scoreboard. Based on the accuracy, the best single model performance is obtained with \camembertUnfrozen{}.
However, a McNemar normalize test \cite{mcnemar1947note} using the contingency table illustrated in \autoref{tab:contigency} of both unfrozen models using only the best seed model per approach (i. e. 20 and 25, respectively) yielded a p-value of $0.5334$. Thus, we can only assume no significant difference between the predictive models.

\begin{table}
    \begin{tabular}{cc|cc}
    \toprule
    \multicolumn{2}{c}{\multirow{2}{*}{}}                                                                & \multicolumn{2}{c}{\textbf{CamemBERT unfrozen}} \\
    \multicolumn{2}{c}{}                                                                                 & Correct           & Incorrect          \\
    \midrule
    \multirow{2}{*}{\begin{tabular}[c]{@{}c@{}}\textbf{CamemBERT} \\ \textbf{unfrozen warmup}\end{tabular}} & Correct   & 1367              & 78                 \\
                                                                                             & Incorrect & 87                & 94     \\
                                                                                             \bottomrule           
    \end{tabular}
    \caption{Contingency table for unfrozen models (token-wise) using only the best seed model per approach (i. e. 20 and 25, respectively)}
    \label{tab:contigency}
    %\vspace{-1em}
\end{table}

\begin{figure}
     \centering
     \captionsetup{width=\linewidth}
     \raisebox{-10ex}
{
    \begin{minipage}{.5\textwidth}
      \centering
        \includegraphics[width=\textwidth]{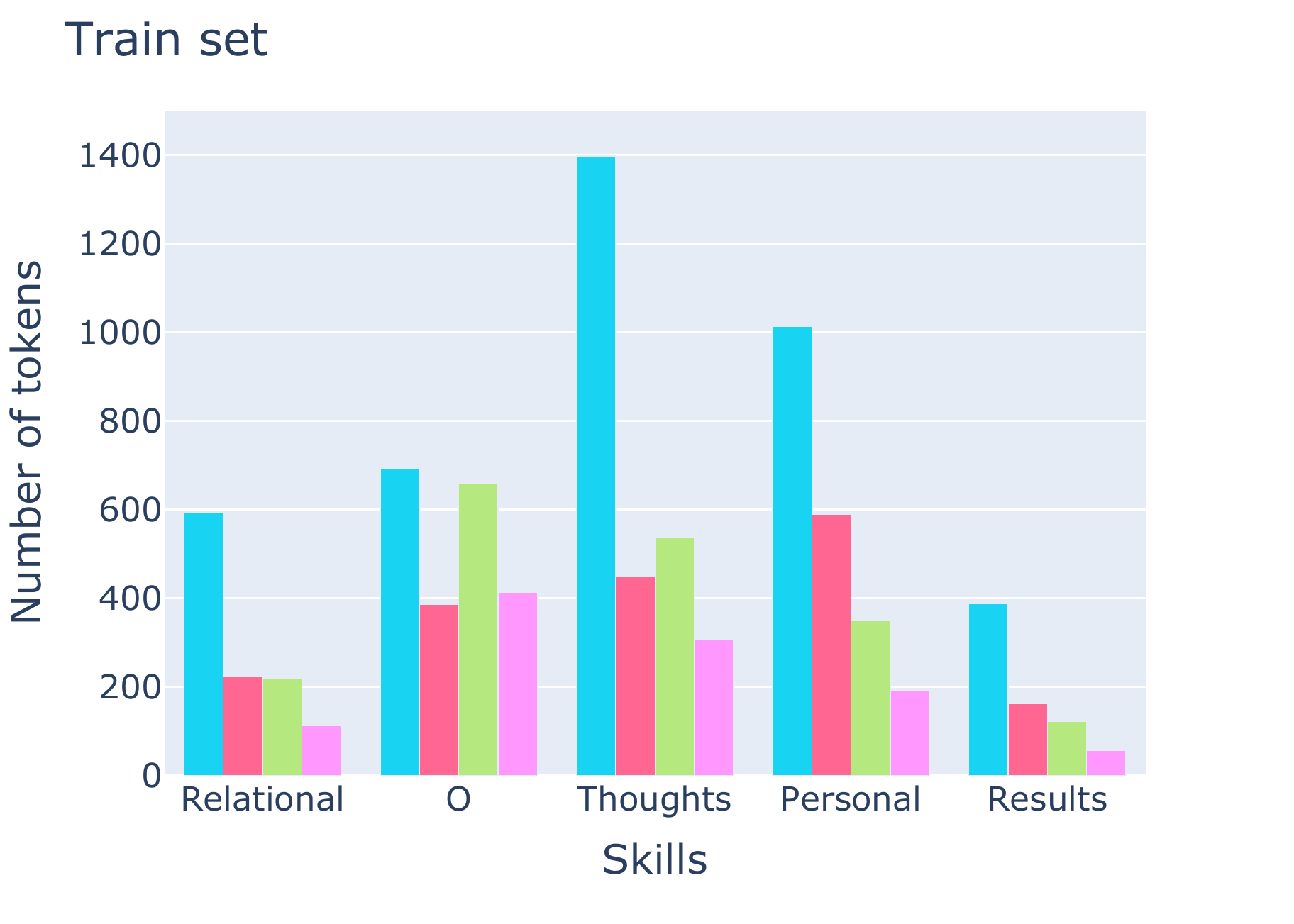}
     \end{minipage}
    \begin{minipage}{.5\textwidth}
      \centering
         \centering
        \includegraphics[width=\textwidth]{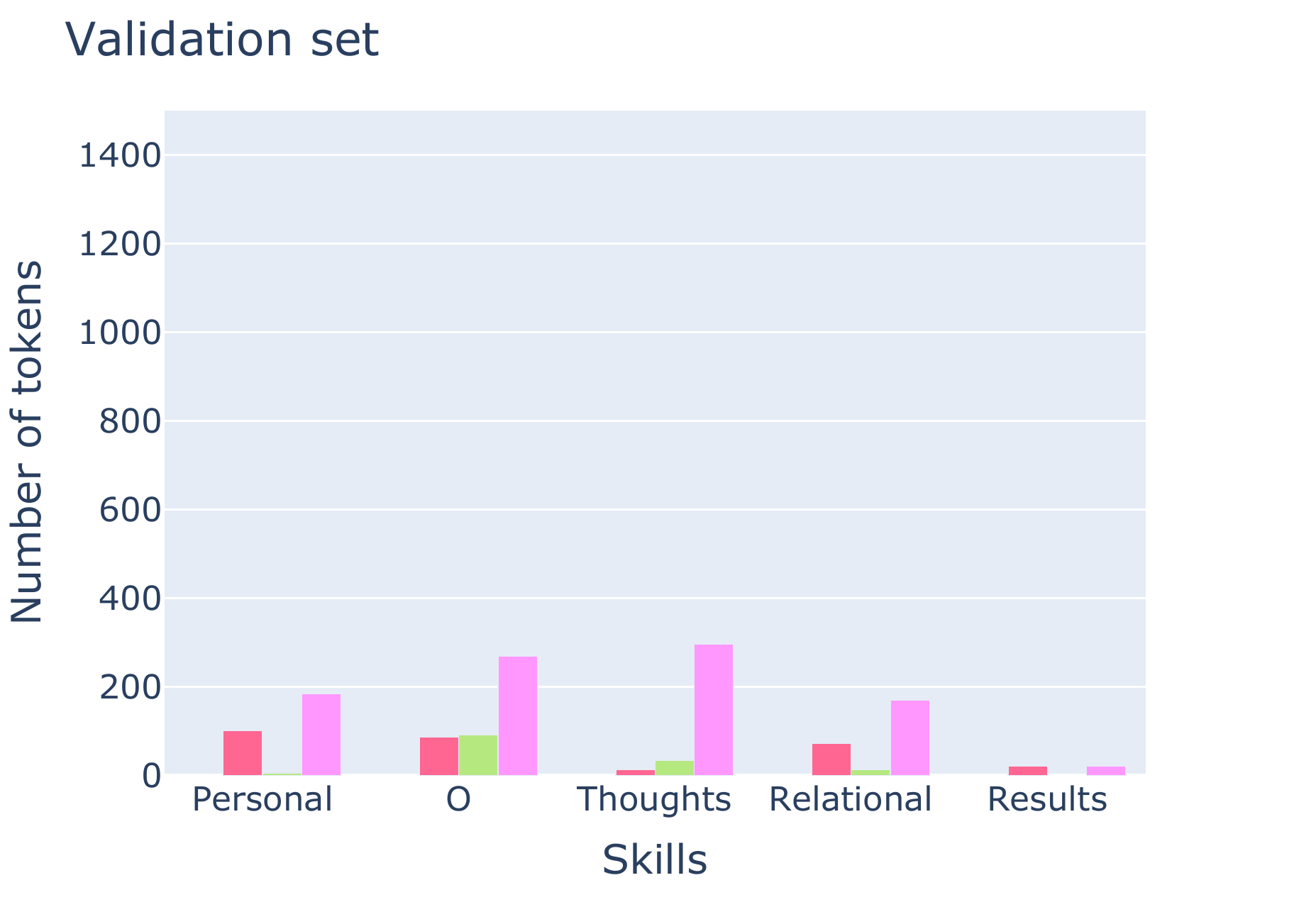}
     \end{minipage}}
     
    \begin{minipage}{.5\textwidth}
      \centering
         \centering
        \includegraphics[width=\textwidth]{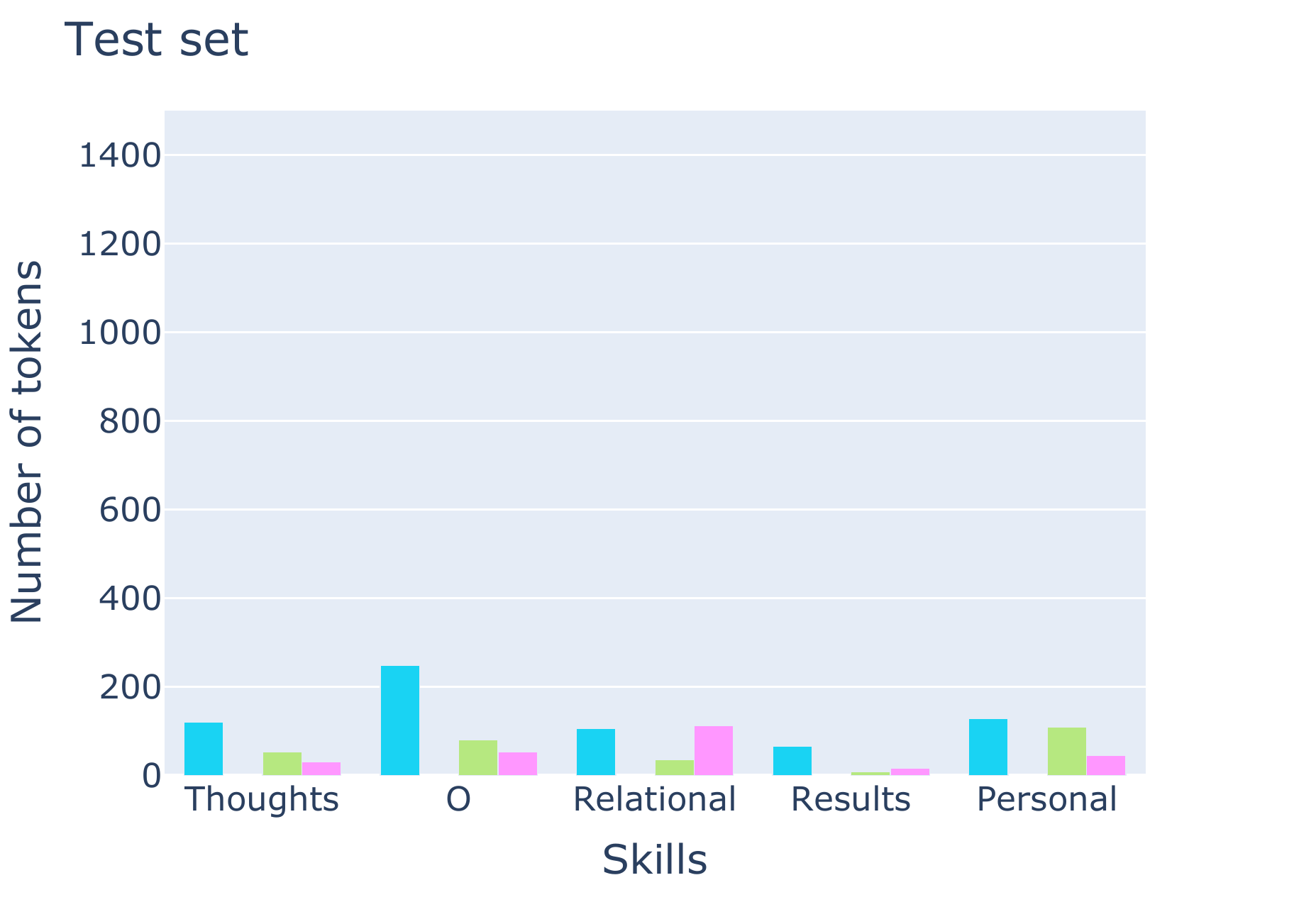}
     \end{minipage}
     \caption{Number of tokens per skill in the annotated dataset (the \guillemet{O} tag means that a token doesn't belong to a skill) where each colour represents one of our four company partners}
    \vspace{-1em}
    \label{fig:split_stats}
\end{figure}

\begin{figure}
    \centering
    \captionsetup{width=\linewidth}
    \includegraphics[width=1.1\textwidth]{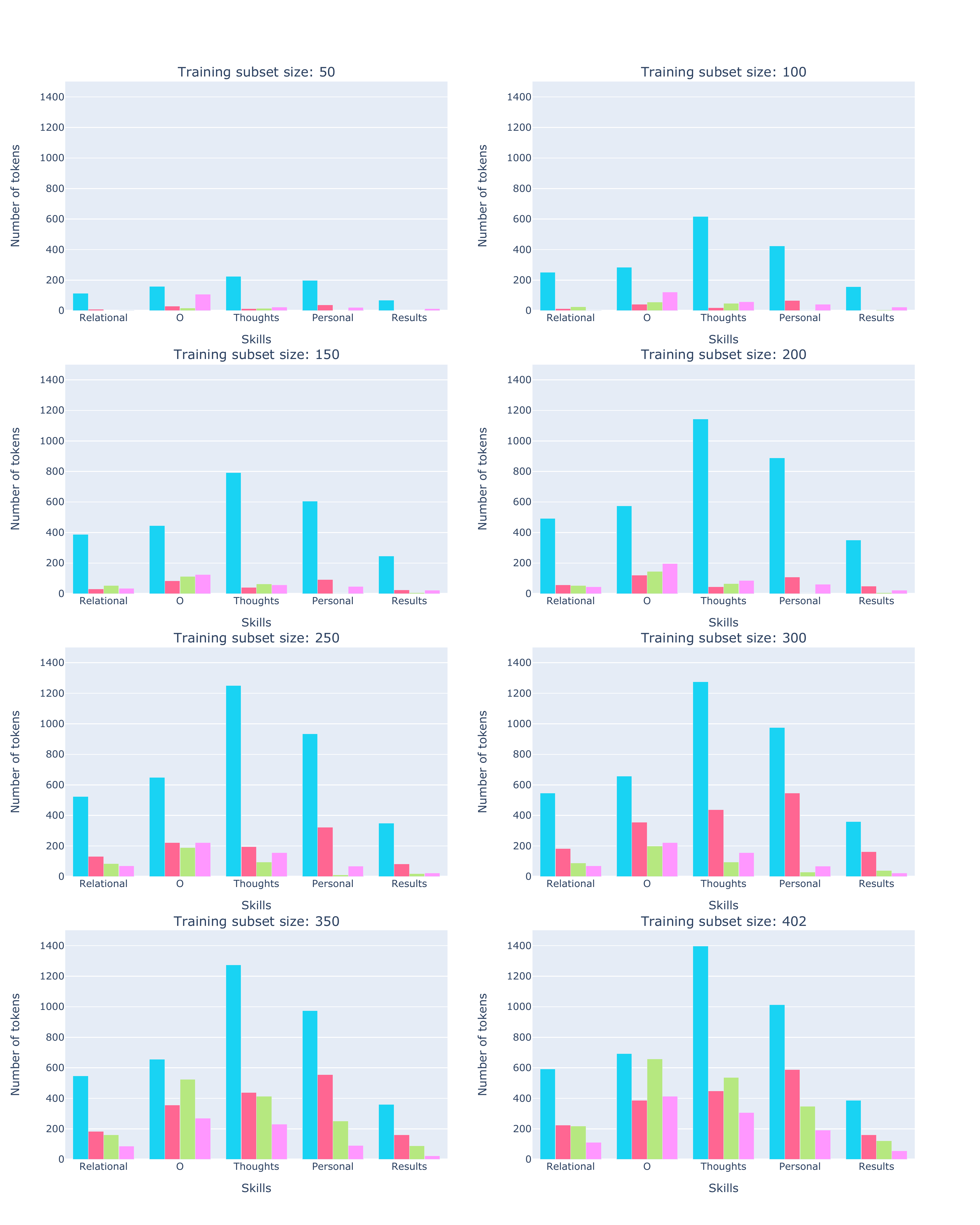}
    \caption{Number of tokens per skill in the training subsets dataset where each colour represents one of our four company partners}
    \label{fig:subset_stats}
\end{figure}

Moreover, the \camembertUnfrozen{} model seems to suffer from a certain degree of instability, as shown by the consistently high standard deviation. This issue mostly persists when using a learning rate warmup followed by a linear decay as proposed by \cite{mosbach2021on}. \camembertFrozen{} is the least sensitive to random initialization while \biLstm{} presents the highest sensitivity and the lowest performance. When it comes to training subsets, we can observe that all models perform best with a high amount of data. However, performance seems quite close across the 200 to 350 subset size range. We hypothesize that this is due to the data distribution of the training and test sets since companies use slightly different ways to express skills. For example, one uses a bullet point style to enumerate skills in a pragmatic approach, while another uses a more situational approach that puts skill within context. Thus, Figure \ref{fig:split_stats} shows that both the train and test sets contain skills belonging, in the majority, to one company (\skyblueemph{sky blue}). Two more companies (\sulueemph{light green} and \violetemph{pink}) are present in the test set, while their skills are underrepresented in the training set. Furthermore, \autoref{fig:subset_stats} shows that the dominant company's (\skyblueemph{sky blue}) skills are well represented in most training subsets, including smaller-sized ones, while augmenting the number of training samples mostly adds skills related to the company that is not present at all in the test set (\pinkemph{red}). It means that the training is quite sensitive to how the data is shuffled because of the limited number of data samples. Therefore, more data would need to be annotated to reach a more balanced dataset and use a stratified random sampling instead of the current simple random sampling to reflect imbalance better when splitting the data. %Morevoer, we also experimented two data augmentation approach to increase the dataset size in order to reduce the imbalancement. The first approach was to add sentence with no annotation that are formal sentence (or filler sentence) such as \guillemet{Join us!}. To do so, we used the exemple in the annotated dataset that does not include any annotation and manually searched in the unannotated similar exemple. This procedure was able to increase the dataset by 53 sentences. The second approach was a weakly annotated approach. We first created a skill embeddings using X of each annotated skill and used the cosine similarity to compute the similarities between each annotated skill with all unannotated sentences. Then, for some annotated sentence we selected the 10 most similar sentence and manually annotated either or not the sentence included the same skill and what portion of the sentence was the same skill (the substring). This procedure allowed us to increase the dataset by 194 examples. However, both approach did not yield better

Finally, \autoref{tab:acc_per_class} presents the mean token-wise accuracy and one standard deviation per skill on the test set for \camembertUnfrozenWarmup{} and \camembertUnfrozen{} for a subset size of $400$.
We can see that for both approaches, the best performance occurs in the class with the lowest example (Results). 
We hypothesize that it might be due to the \guillemet{simplicity} of those examples that are shorter than the average with an average length of around seven tokens.
Thus, these examples are possibly more straightforward than the others, leading to an easier classification.
Also, we observe a higher variance on both Thoughts and Personal class for both our models.
These two tags have the highest standard deviation, even if they are the classes with the most examples.
It means that the training is quite sensitive to the initialization of the models. Because we have a limited number of data samples, minimizing such instability during training is more difficult. Therefore, more data would need to be annotated to reach a more stable training for all classes.
Furthermore, we can see that there is still room for improvement for most tags.

\begin{table}
    \centering
    \captionsetup{width=\linewidth}
    \resizebox{\textwidth}{!}{\begin{tabular}{ccccc|ccccc}
\toprule
\multicolumn{5}{c}{\camembertUnfrozenWarmup} & \multicolumn{5}{c}{\camembertUnfrozen} \\
                    O &            Thoughts &         Results &        Relational &         Personal &                    O &             Thoughts &          Results &        Relational &          Personal \\
\midrule
$84.50 \pm 4.51$ &           $82.72 \pm 7.84$ &  $91.21 \pm 0.00$ &          $73.41 \pm 4.44$ &  $80.77 \pm 4.97$ &            $83.30 \pm 3.63$ &          $80.97 \pm 6.56$ &  $92.31 \pm 4.85$ &          $77.73 \pm 3.48$ &  $85.21 \pm 9.65$ \\
\bottomrule
\end{tabular}}
    \caption{Mean token-wise accuracy and one standard deviation per skill for \camembertUnfrozenWarmup{} and \camembertUnfrozen{} on the test set for a subset size of $400$ (the \guillemet{O} tag means that a token does not belong to a skill)}
    \label{tab:acc_per_class}
\end{table}

\subsection{Error Analysis}
Using the approaches that yielded the higher accuracy (\camembertUnfrozen{}), we conducted an error analysis on is 24 errors. We found that most of these were types that are similar to the cases illustrated in \autoref{fig:anot_conf2}, namely two consecutive skill annotations with the same class but separated by a coordination conjunction. The NER identified the two skills as a single skill in all those error cases.
Moreover, some cases (3) were a similar error type where a part of the sentence, before a coordination conjunction, is not an entity as illustrated in \autoref{tab:firstexample}.
The figure introduces each token's ground truth and prediction (\guillemet{Prob} rows), along with the model probabilities, using the same color scheme as \autoref{fig:pca_tf-idf}, namely \rougeemph{red} is the \guillemet{Thoughts} class, \mauveemph{purple} is \guillemet{Personal}, \blueemph{blue} is \guillemet{Relational}, \orangeemph{orange} is \guillemet{Results}, and \vertemph{green} is a word not in an entity.
We can see that not only does the NER wrongly predict the class, Personal rather than Thoughts, but it also wrongly predicted that the first part of the sentence, \guillemet{\textit{les dossiers sont de nature courante}}, is also a skill.
We hypothesize that it is due to two things.
First of all, the presence of the words \guillemet{\textit{dossiers}}, \guillemet{\textit{nature}}, \guillemet{\textit{courante}} that appear in other Personal examples.
For instance, the word \guillemet{\textit{dossiers}} appears 49 times in a Personal entity, which could confuse our model as to whether such a sentence piece is a skill.
Second of all, the presence of the coordinating conjunction \guillemet{\textit{et}} plus the determinant \guillemet{\textit{les}} which mostly appear within an entity and rarely appear just outside of it (a token before).
We argue that our model annotated the overall sentence as a skill in that specific case due to the overwhelming examples of such cases.
However, the entity class prediction is inconsistent with the sentence vocabulary distribution.
The second part of the sentence is composed of words that only appear in Thoughts examples, such as \guillemet{\textit{décision}}, \guillemet{\textit{analyse}} and \guillemet{\textit{recherche}}.
It leads to lower probability confidence of our NER, where those three words have the lowest probabilities.

\begin{figure}
    \captionsetup{width=\linewidth}
    \centering
    \begin{tabular}{l|ccccccccc}
    \toprule
    Token       & \vertemph{les}        & \vertemph{dossiers} & \vertemph{sont}     & \vertemph{de}       & \vertemph{nature}   & \vertemph{courante} & \vertemph{et}       & \rougeemph{les}      & \rougeemph{décisions} \\
    Prob & \mauveemph{0.82}       & \mauveemph{0.80}     & \mauveemph{0.83}     & \mauveemph{0.84}     & \mauveemph{0.83}     & \mauveemph{0.83}     & \mauveemph{0.78}     & \mauveemph{0.53}     & \mauveemph{0.49}      \\\midrule
    Token       & \rougeemph{requièrent} & \rougeemph{un}       & \rougeemph{niveau}   & \rougeemph{habituel} & \rougeemph{d'}       & \rougeemph{analyse}  & \rougeemph{et}       & \rougeemph{de}       & \rougeemph{recherche} \\
    Prob & \mauveemph{0.81}       & \mauveemph{0.85}     & \mauveemph{0.82}     & \mauveemph{0.83}     & \mauveemph{0.78}     & \mauveemph{0.51}     & \mauveemph{0.73}     & \mauveemph{0.74}     & \mauveemph{0.64}     \\\bottomrule
    \end{tabular}
    \caption{Example of a wrongly predicted sentence using the best seed \camembertUnfrozen{} model where color represent the skill class (\rougeemph{red} is the \guillemet{Thoughts} class, \mauveemph{purple} is \guillemet{Personal}, \blueemph{blue} is \guillemet{Relational}, \orangeemph{orange} is \guillemet{Results}, and \vertemph{green} is a word not in an entity)}
    \label{tab:firstexample}
\end{figure}

%For our analysis, we conducted a quantitative token class prediction error and a qualitative error of some selected errors.

%First, we present in \autoref{tab:top5} the top 5 most misclassification proportion errors type, where the first element is the NER predicted entity type and the second element is the ground truth.
%The number of errors have been weighed by the ground truth number of tokens, since we have an unbalanced number of entity.
%First, we can see that despite the clear centroid distinction shown in \autoref{fig:pca_tf-idf} between the two classes \guillemet{Thoughts} and \guillemet{Personal}, it appear than some tokens are quite confusing thus 

%Thoughts and Personal are the two leading class in number of entities.

%\begin{table}
%    \begin{tabular}{lc}
%    \toprule
%    Misclassification     & \begin{tabular}[c]{@{}c@{}}Proportional Number of Error (\%)\\($\frac{\texttt{\# of Errors}}{\texttt{\# Ground Truth Tokens}}$)\end{tabular}\\\midrule
%    (Thoughts, Relational)   & 2.26              \\
%    (Personal, Out-of-Entity) & 1.21              \\
%    (Personal, Thoughts)        & 1.04              \\
%    (Out-of-Entity, Relational)           & 0.78              \\
%    (Out-of-Entity, Thoughts) & 0.67      \\\bottomrule       
%    \end{tabular}
%    \caption{texte}
%    \label{tab:top5}
%\end{table}

\section{Conclusion}
\label{sec:conclusion}

This article presents a new public dataset in French, including annotated and non-annotated job offers in the insurance domain. It aims to develop machine learning models to perform automatic skill recognition inside job ads, which is an increasingly useful task for understanding the evolution of the labour market. The dataset statistics and characteristics show limitations that could have made it challenging to perform the learning task well. It can be further improved by rebalancing companies' and skill classes' to make the annotated dataset more representative of the non-annotated distribution.  Moreover, the impact of lexical overlapping and soft skill identification could be lowered by allowing more experts to annotate more job ads. In any case, this dataset will be improved by adding annotations.

Despite these dataset limitations, we have obtained interesting results with pre-trained models despite the size of the dataset with a token-wise approach. Although the skill-wise problem is closer to our main objective, our preliminary experiments on the skill-wise problem with common NLP algorithms seem to lead to poor accuracy. Since our results here are token-wise and not skill-wise, it is harder to extract the correct span of skill entities, and consequently, we cannot conclude the actual number of skills inside the non-annotated dataset. Thus, our work is a first step toward discovering some trends in the labour market and studying the evolution of skills. As our next step, our objective is to make efficient models identifying skills instead of tokens. Indeed, our models cannot distinguish two skills with the same tag if they are next to each other. Therefore, we need to detect the beginning of each skill inside the text. To achieve that, we plan to use the BIO tags scheme instead of IO \cite{konkol2015segment}. However, adding a new beginning tag for each skill group would probably reduce the overall accuracy because of the small size of the dataset. At last, improving the cleaning phase by detecting more precisely the usual conjunctions between two skills could be another way to keep the token-wise results and identify the skills more efficiently.

We have made some error analyses, but this kind of analysis is limited by the limited explainability possibilities of deep learning models. In the long term, we would like to make our model more explainable both to help us understand the strengths and weaknesses of our model and explain results to recruiters and human resources staff to allow them to adapt their needs in recruitment. To do so, we plan to explore counterfactual generation \cite{madaan2021generate, fern2021text}.

Finally, we have only explored a few possibilities this dataset can offer. Some other tasks that can be performed could include the measurement of the impact of redaction style (e. g. long sentences vs. bullet points, different ways of addressing a potential applicant) on the performance, the impact of gendered wording \cite{gaucher2011evidence}, the impact of COVID and teleworking on skill requirements \cite{gaffney2021trends} or even the extension of the dataset to include new tools used by human resources such as social media \cite{ruparel2020influence}.

\section*{Acknowledgements}
This research was made possible thanks to the support of the Future Skills Centre and four Canadian insurance companies. We wish to thank the reviewers for their comments regarding our work and methodology.

\printbibliography[heading=subbibintoc]

\end{document}